\documentclass{article}

\usepackage{lineno}
\usepackage{booktabs}
\usepackage{authblk}
\usepackage{multirow}
\usepackage{graphicx}
\usepackage[numbers]{natbib}
\usepackage{hyperref}
\usepackage{fixltx2e}
\usepackage{subcaption}
\usepackage[preprint]{tackling_climate_workshop_style}

\usepackage[utf8]{inputenc} % allow utf-8 input
\usepackage[T1]{fontenc}    % use 8-bit T1 fonts
\usepackage{hyperref}       % hyperlinks
\usepackage{url}            % simple URL typesetting
\usepackage{booktabs}       % professional-quality tables
\usepackage{amsfonts}       % blackboard math symbols
\usepackage{nicefrac}       % compact symbols for 1/2, etc.
\usepackage{microtype}      % microtypography

\title{\textit{} Paraformer: Parameterization of Sub-grid Scale Processes Using Transformers}

\author[1]{\textbf{Shuochen Wang}}
\author[2]{\textbf{Nishant Yadav}}
\author[1, 3, 4, 5]{\textbf{Auroop R. Ganguly}}
\affil[1]{Sustainability and Data Sciences Laboratory, Northeastern University}
\affil[2]{Microsoft Corporation}
\affil[3]{The Institute for Experiential AI and Roux Institute, Northeastern University}
\affil[4]{Zeus AI}
\affil[5]{Pacific Northwest National Laboratory}

\begin{document}

\maketitle

\begin{abstract}
One of the major sources of uncertainty in the current generation of Global Climate Models (GCMs) is the representation of sub-grid scale physical processes. Over the years, a series of deep-learning-based parameterization schemes have been developed and tested on both idealized and real-geography GCMs. However, datasets on which previous deep-learning models were trained either contain limited variables or have low spatial-temporal coverage, which can not fully simulate the parameterization process. Additionally, these schemes rely on classical architectures while the latest attention mechanism used in Transformer models remains unexplored in this field. In this paper, we propose Paraformer, a “memory-aware” Transformer-based model on ClimSim, the largest dataset ever created for climate parameterization. Our results demonstrate that the proposed model successfully captures the complex non-linear dependencies in the sub-grid scale variables and outperforms classical deep-learning architectures. This work highlights the applicability of the attention mechanism in this field and provides valuable insights for developing future deep-learning-based climate parameterization schemes.
\end{abstract}

\section{Introduction}

Understanding and modeling complex physical processes in the Earth system are crucial for making prompt and precise decisions regarding climate change issues. The Global Climate Model (GCM) is a tool to simulate past climate changes and provide potential trajectories for the future, accounting for both natural and anthropogenic activities. It achieves this by modeling critical physical processes in various components of the Earth system, including, but not limited to, the atmosphere, ocean, land, and cryosphere. Since the late 20th century, GCMs have advanced significantly through joint efforts from Earth system scientists and computer scientists. With the aid of supercomputers, substantial improvements in model complexity, spatial-temporal resolution and prediction accuracy have been realized in the current generation of GCMs compared to their earlier counterparts. However, the latest GCMs integrated by the Coupled Model Intercomparison Project Phase 6 (CMIP6) still feature a coarse spatial resolution of 100–250 km and exhibit a wide spread in simulation results, primarily due to challenges associated with sub-grid-scale physical processes. For instance, research has shown that the representation of small-scale clouds is critical in a general circulation model \citep{senior1993carbon,slingo1989gcm} and it is still one of the major sources of uncertainty in the current GCMs \citep{schneider2017climate,zelinka2020causes}. Other studies have investigated the impacts of sub-grid hydrologic processes \citep{thomas1991evaluation,hagemann2003improving}, turbulent flows \citep{yano2016subgrid}, radiative transfer \citep{belochitski2021robustness,yao2023physics} and topography \citep{im2010validation} on model simulation. Despite the advancements in GCMs, these sub-grid scale processes cannot be explicitly resolved by the model and thus must be represented by approximations using the grid-scale processes, a technique known as parameterization.

Over the years, a series of parameterization schemes have been developed and applied to weather forecasting \citep{whitaker2008ensemble,grell2002generalized} and climate prediction \citep{del1996prognostic,jakob2003improved,pedatella2014ensemble} using traditional data assimilation methods. These methods integrate observational data with numerical models to estimate the state of the Earth’s system. However, data assimilation is subject to biases arising from the assimilating model, the quality of observations, and the method itself \citep{dee2005bias}. Additionally, incorporating vast amounts of observational data into forecasting systems is computationally expensive \citep{howard2024machine}. In recent years, there has been a growing interest in developing parameterization schemes using data-driven approaches like Deep Learning (DL) \citep{rasp2018deep,gentine2018could}. A significant challenge with this approach, however, is the acquisition of sub-grid scale information since it requires high-resolution model runs. Some studies have taken an alternative approach by testing parameterization schemes on highly idealized models such as the Lorenz 96 system, which comprises two sets of differential equations representing changes in grid-scale and sub-grid scale variables \citep{wilks2005effects,gagne2020machine,arnold2013stochastic}. While this approach is useful, testing parameterization schemes directly on real-geography GCMs is preferable since they are closer to the real climate we know. However, most of the datasets previously used to develop DL-based parameterization schemes either included limited variables or offered low spatial-temporal coverage, hindering the application of more advanced architectures. Recently, the newly published ClimSim dataset addressed this challenge by providing the largest and most physically comprehensive testbed for DL parameterization schemes on climate simulations \citep{yu2024climsim}.

In recent years, various DL architectures have been tested on weather and climate parameterization problems. Current DL-based schemes primarily use classical architectures such as Random Forest \citep{o2018using,yuval2020stable}, Convolutional Neural Networks \citep{hu2024stable,bolton2019applications,larraondo2019data}, Multi-Layer Perceptron \cite{gentine2018could,song2021improved,krasnopolsky2013using}, and Generative Adversarial Networks \citep{gagne2020machine,nadiga2022stochastic,perezhogin2023generative}. These architectures are chosen primarily due to their relatively low computational costs for training \citep{song2021improved}. In 2017, a novel DL architecture called Transformer was introduced in Natural Language Processing \citep{vaswani2017attention}. It was soon applied to related tasks such as machine translation \citep{wang2019learning} and text generation \citep{li2024pre}. Since then, Transformer architecture has gained significant popularity across a wide range of disciplines, including computer science and beyond. The attention mechanism in Transformers performs well at handling long-range dependencies in sequential data and thus has been used for time series prediction recently \citep{zhou2021informer,zerveas2021transformer,wen2022transformers}. Despite their success in other fields, Transformers have not yet been investigated for climate parameterization. The attention mechanism allows Transformer models to focus on specific parts of the input data, assigning importance scores to previous data points. In this way, a Transformer model can generate a dynamic memory bank while predicting the next item. For parameterization schemes, it translates to the model capturing inter-state and intra-state temporal dependencies while approximating sub-grid scale processes.

In this work, we propose a Transformer-based “memory-aware” parameterization scheme (hereafter called Paraformer) using the ClimSim dataset, introducing the attention mechanism to this domain. Our results demonstrate that the attention mechanism effectively captures the complex nonlinear dependencies of sub-grid-scale processes in the Earth's system, achieving lower prediction errors compared to classical DL architectures.

\section{Data and Methods}

\subsection{The ClimSim Dataset}

ClimSim is the largest-ever dataset for developing machine-learning emulators on climate parameterization published in 2023 \citep{yu2024climsim}. This dataset includes a variety of input and output variables generated by the Energy Exascale Earth System Model (E3SM) with a Multiscale Modeling Framework (MMF), representing the grid-scale and sub-grid scale physical processes in the Earth's system. The MMF replaces the traditional parameterization scheme with a Cloud Resolving Model (CRM), which explicitly resolves small-scale physics such as clouds and turbulence. Several classical DL architectures were implemented as baselines in the original study \citep{yu2024climsim}.

In this work, we utilize the low-resolution, real-geography dataset from the ClimSim suite, which contains 10 years of data at 384 unstructured spatial grids with a temporal resolution of 20 minutes. The dataset has a total size of 744 GB with 10,099,200 data points. We subsample the data to have an effective temporal resolution of 140 minutes while keeping the spatial structure intact. We conduct two series of experiments: one is based on a subset of critical input and output variables (v1) and the other uses an expanded set of variables (v2). v1 includes 124 input and 128 output variables, and v2 includes 557 input and 368 output variables. Details of these variables are provided in Table \ref{tab:v1} and \ref{tab:v2}. For both v1 and v2, we split the dataset into training, validation, and test sets. The training set spans 7 years (0001-02 to 0008-01) and the validation set covers 1 year (0008-02 to 0009-01). For v1, the test set has the same temporal coverage as the validation set but with a temporal resolution of 120 minutes to approximate a different climate model run. For v2, the test set is from the actual test data covering the remaining 2 years (0009-03 to 0011-02). The temporal subsampling and variable selection serve two purposes: (1) addressing memory limitations that prevent using the entire dataset without subsampling, and (2) ensuring consistency with the original ClimSim paper \citep{yu2024climsim}, facilitating performance comparisons with the baseline models.

\subsection{The Architecture of Paraformer}

\begin{figure}[h]
    \centering
    \includegraphics[width=0.83\linewidth]{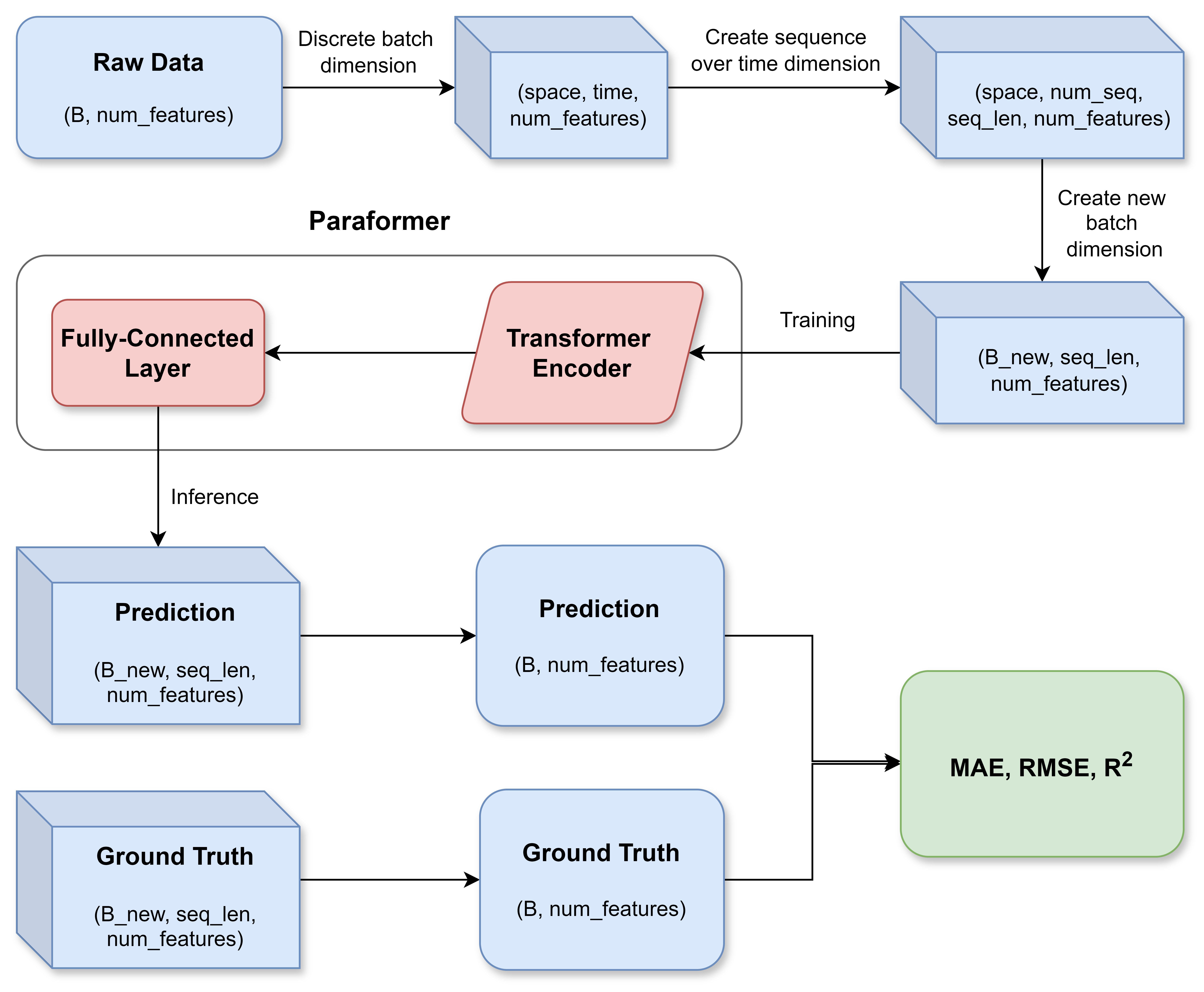}
    \caption{The architecture of Paraformer and the data processing workflow. The shape of data is labeled in parentheses for each step. \texttt{B} and \texttt{B\textunderscore new} represent two different batch dimensions of data. In the raw data, \texttt{B} combines the spatial and temporal dimensions. \texttt{num\textunderscore features} refers to the number of input (grid-scale) and output (sub-grid scale) variables. \texttt{num\textunderscore seq} and \texttt{seq\textunderscore len} represent the number and length of sequences, respectively. Metrics for prediction accuracy include Mean Absolute Error (MAE), Root Mean Square Error (RMSE) and Coefficient of Determination (R\textsuperscript{2}).}
    \label{fig:architecture}
\end{figure}

Figure \ref{fig:architecture} demonstrates the details of the architecture and data processing workflow. The architecture of Paraformer consists of an encoder-only Transformer \citep{vaswani2017attention} followed by a fully-connected layer. The model is defined by its context window size (i.e. the number of sequential inputs it processes in a single pass). The context window size controls how much memory the model holds at a time. In ClimSim, we generate the context of a set size (hyperparameter) by creating sequential windows over the time dimension. We tested two approaches: the first involves creating a series of sliding windows that move one step at a time across the time series. This approach is analogous to a text generation problem, where a word or letter is generated based on the preceding sequence of text. We tested different window sizes ranging from 5 to 40. However, one issue with this method is data duplication, which significantly increases the dataset size depending on the selected window size, thereby increasing training time. Therefore, we tested a second method, where the time dimension is split into independent, non-overlapping windows. We found a window size of 5 generated the best results and creating a sliding window does not outperform the simpler non-overlapping window approach. A window size of 5 is equal to roughly 12 hours of climate memory in the model at the given temporal resolution.

\subsection{Hyperparameter Search}

% \usepackage{booktabs}
% \usepackage{graphicx}
\begin{table}[h]
\centering
\caption{Details of the hyperparameter search process.}
\label{tab:hyperparam}
\resizebox{0.8\textwidth}{!}{%
\begin{tabular}{@{}ll@{}}
\toprule
\textbf{Hyperparameter}            & \textbf{Domain}                                     \\ \midrule
Number of Encoder Layers  & {[}2, 4, 6, 8, 10, 12{]}                   \\
Embedding Dimension       & {[}64, 128, 256, 512{]}                    \\
Number of Attention Heads & {[}4, 8{]}                                 \\
Batch Size & {[}64, 128, 256, 512{]}                                 \\
Optimizer                 & {[}SGD, Adam, AdamW{]}                     \\
Learning Rate Scheduler   & {[}CosineAnnealingLR, ReduceLROnPlateau{]} \\ \bottomrule
\end{tabular}%
}
\end{table}

In this work, we tested different combinations of hyperparameters in a grid search process summarized in Table \ref{tab:hyperparam}. We performed the hyperparameter search using multiple nodes on a high-performance computing cluster, with one NVIDIA A100 GPU per node.

\section{Results}

% We performed a hyperparameter search process summarized in \ref{appendix_b}. 
We performed the hyperparameter search process for v1 and v2 independently. The best hyperparameter configuration for v1 is: an embedding dimension of 256, 6 Transformer encoder layers, 4 attention heads, a batch size of 512, and AdamW optimizer. The model’s loss function is taken as mean square error (MSE) and the learning rate is defined using a ReduceLROnPlateau learning rate scheduler with a patience of 10 epochs, a reduction factor of 0.5 for a total of 200 epochs, and an initial learning rate of 1$\times$10\textsuperscript{-4}. For v2, the embedding dimension increases to 512 to accommodate the expanded feature set, while all other hyperparameters remain the same. To establish a baseline, we reimplemented the Multi-Layer Perceptron (MLP) model based on the configuration provided in \citep{yu2024climsim}. The MLP was chosen as it is one of the most commonly used DL architectures for climate parameterization and demonstrated the highest overall prediction accuracy for most output variables in \citep{yu2024climsim}. It is important to note that the hyperparameters of the MLP differ between v1 and v2, reflecting the independent search processes conducted.

% \usepackage{booktabs}
% \usepackage{multirow}
% \usepackage{graphicx}
\begin{table}[h]
\centering
\caption{MAE and R\textsuperscript{2} for v1 using the six baseline models provided in \citep{yu2024climsim} and the Paraformer model. Metrics for MLP are from the reimplemented version while values for other baselines are identical to \citep{yu2024climsim}. Large negative global R\textsuperscript{2} are not shown for $dq/dt$ and PRECSC due to the variabilities in the upper atmosphere and tropics, respectively. Units of non-energy flux variables are converted to a common energy unit, W/m\textsuperscript{2} \citep{yu2024climsim}. The best model for each variable is bolded.}
\label{tab:1}
\resizebox{\textwidth}{!}{%
\begin{tabular}{@{}lllllllllllllll@{}}
\toprule
\multicolumn{1}{c}{\multirow{2}{*}{\textbf{Variables}}} & \multicolumn{7}{c}{\textbf{MAE {[}W/m\textsuperscript{2}{]}}}                  & \multicolumn{7}{c}{\textbf{R\textsuperscript{2}}}                                   \\ \cmidrule(l){2-15} 
\multicolumn{1}{c}{}                                    & CNN   & ED     & HSR   & MLP   & RPN   & cVAE  & Paraformer & CNN   & ED      & HSR    & MLP    & RPN    & cVAE   & Paraformer \\ \cmidrule(r){1-15}
$dT/dt$                                                  & 2.585 & 2.684  & 2.845 & 2.673 & 2.685 & 2.732 & \textbf{2.332}           & 0.627 & 0.542   & 0.568  & 0.594  & 0.617  & 0.59   & \textbf{0.681}           \\
$dq/dt$                                                   & 4.401 & 4.673  & 4.784 & 4.519 & 4.592 & 4.68  & \textbf{4.049}           & -     & -       & -      & -      & -      & -      & -           \\
NETSW                                                   & 18.85 & 14.968 & 19.82 & 13.753 & 18.88 & 19.73 & \textbf{9.739}           & 0.944 & 0.98    & 0.959  & 0.982  & 0.968  & 0.957  & \textbf{0.990}           \\
FLWDS                                                   & 8.598 & 6.894  & 6.267 & 5.410 & 6.018 & 6.588 & \textbf{4.471}           & 0.828 & 0.802   & 0.904  & 0.917  & 0.912  & 0.883  & \textbf{0.940}           \\
PRECSC                                                  & 3.364 & 3.046  & 3.511 & 2.687 & 3.328 & 3.322 & \textbf{2.285}           & -     & -       & -      & -      & -      & -      & -           \\
PRECC                                                   & 37.83 & 37.25  & 42.38 & 33.838 & 37.46 & 38.81 & \textbf{22.199}           & \textbf{0.077} & -17.909 & -68.35 & -34.545 & -67.94 & -0.926 & -1.764           \\
SOLS                                                    & 10.83 & 8.554  & 11.31 & 8.163  & 10.36 & 10.94 & \textbf{6.529}           & 0.927 & 0.96    & 0.929  & 0.959  & 0.943  & 0.929  & \textbf{0.972}           \\
SOLL                                                    & 13.15 & 10.924 & 13.6  & 10.562  & 12.96 & 13.46 & \textbf{8.950}           & 0.916 & 0.945   & 0.916  & 0.945  & 0.928  & 0.915  & \textbf{0.958}           \\
SOLSD                                                   & 5.817 & 5.075  & 6.331 & 4.603 & 5.846 & 6.159 & \textbf{3.701}           & 0.927 & 0.951   & 0.923  & 0.955  & 0.94   & 0.921  & \textbf{0.969}           \\
SOLLD                                                   & 5.679 & 5.136  & 6.215 & 4.841 & 5.702 & 6.066 & \textbf{4.318}           & 0.813 & 0.857   & 0.797  & 0.863  & 0.837  & 0.796  & \textbf{0.888}           \\ \bottomrule
\end{tabular}%
}
\end{table}

Table \ref{tab:1} shows MAE and R\textsuperscript{2} values for the target sub-grid scale variables in v1 using Paraformer and other six baseline models provided in \citep{yu2024climsim}. The Paraformer model outperforms all other structures on all but one variable, indicating that the attention mechanism effectively captures the complex temporal dependencies of the sub-grid scale physical processes. Metrics for v2 are shown in Table \ref{tab:metric v2}. Paraformer demonstrates lower errors on all variables in v2 compared to the MLP baseline. However, the improvement is relatively modest, suggesting that parameterizing more complex physical processes in the Earth system may require even more advanced DL models.

\begin{figure}[htbp]
    \centering
    \includegraphics[width=1\linewidth]{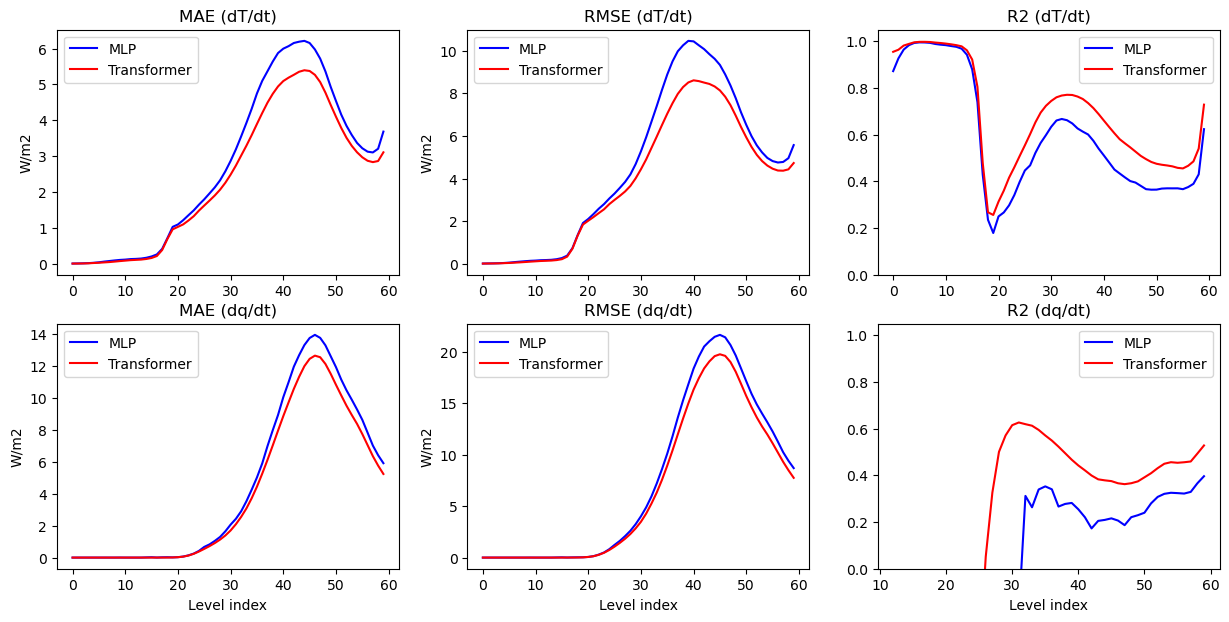}
    \caption{MAE, RMSE and R\textsuperscript{2} of $dT/dt$ and $dq/dt$ using MLP and Paraformer on variable set v1. Each index on the x-axis represents a vertical level in the atmosphere starting from the top (i.e. level index 0 represents the top of the atmosphere). Units of non-energy flux variables are converted to a common energy unit, W/m\textsuperscript{2} \citep{yu2024climsim}. Negative R\textsuperscript{2} values are not shown.}
    \label{fig:1}
\end{figure}

\begin{figure}[htbp]
    \centering
    \includegraphics[scale=0.25]{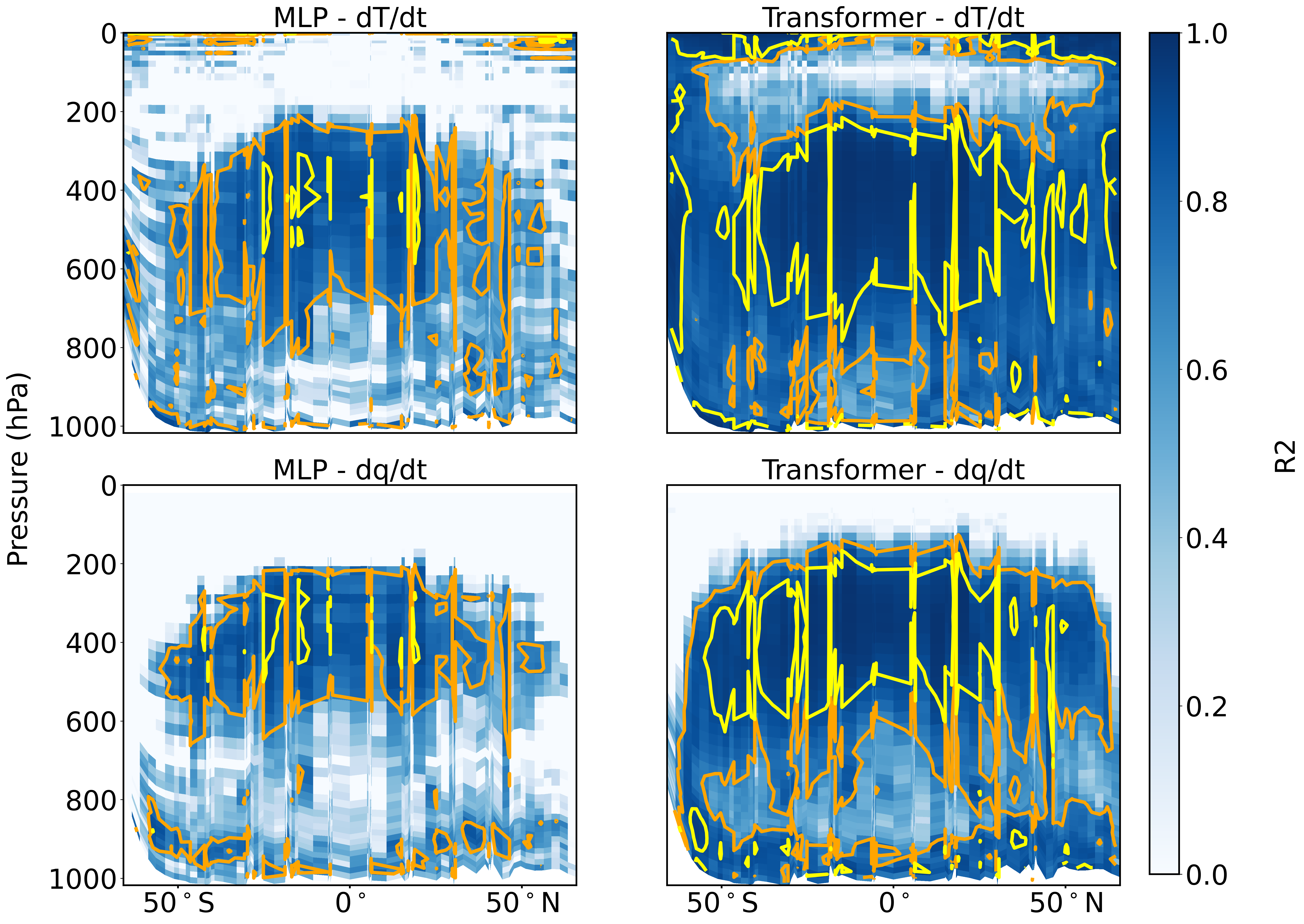}
    \caption{R\textsuperscript{2} of daily-mean, zonal-mean $dT/dt$ and $dq/dt$ for MLP and Paraformer at different pressure levels in variable set v1. Yellow contours cover regions of > 0.9R\textsuperscript{2}, orange contours cover regions of > 0.7R\textsuperscript{2}.}
    \label{fig:2}
\end{figure}

To visualize the prediction improvement of variables with vertical structures, we plotted the metrics for $dT/dt$ and $dq/dt$ and compared them to the MLP model. We also analyzed them for v2, as these are the only two variables with vertical structures shared by both sets. As shown in Figure \ref{fig:1}, in the upper levels of the atmosphere (level index 0 to 20), both models exhibit similar performance for $dT/dt$ and $dq/dt$, However, performance diverges at lower levels, with Paraformer achieving significant reductions in MAE and RMSE at level indices around 40. Paraformer also shows a general improvement in R\textsuperscript{2} across atmospheric levels 20 to 60 for $dT/dt$ and 30 to 60 for $dq/dt$. For v2 (Figure \ref{fig1 v2}), increasing the number of variables improves prediction accuracy for both models on $dT/dt$ and $dq/dt$, with a larger magnitude of improvement for MLP than for Paraformer. Notably, Paraformer produces a smoother R\textsuperscript{2} across different atmospheric levels compared to the MLP. Figure \ref{fig1 v2 cloud} illustrates the prediction performance for two output variables related to cloud tendencies in v2. As previously noted, the Paraformer model shows limited improvement. For wind tendencies ($du/dt$ and $dv/dt$), no significant differences in MAE or RMSE are observed between the two models, as these values are negligible. In Figures \ref{scatter2} and \ref{scatter3}, we compare model predictions to the ground truth of all data points on selected atmospheric levels. Many data points have values close to zero across all levels, a characteristic that both models capture well. For $dT/dt$, in the upper levels (e.g., level 1), most data points exhibit a small magnitude, which both models can predict accurately. In contrast, for $dq/dt$, since nearly all data points have small values at these levels, the prediction accuracy is low for both models, showing large negative R\textsuperscript{2}. The Paraformer model improves accuracy for uncommon data points (those with values far from zero), producing a higher R\textsuperscript{2} value than MLP. At lower levels (e.g., levels 40 and 59), Paraformer effectively captures positive and negative trends. However, at intermediate levels (e.g., level 25), both models struggle to capture certain trends. In general, the Paraformer model improves the prediction accuracy of non-scalar variables across the atmosphere, though the extent of improvement varies.

In Figure \ref{fig:2}, we analyze prediction performance across different pressure levels and latitudes. Both models demonstrate higher precision in the low-latitude 400-hPa region, whereas R\textsuperscript{2} values decrease around 50\textdegree S and 50\textdegree N. This decline may result from the ClimSim dataset’s structure, as its spatial grids are unevenly distributed, with fewer grids available near the poles compared to the equatorial region. Consistent with Figure \ref{fig:1}, the Paraformer model generally enhances the overall accuracy but shows limited improvement in the upper atmosphere (0 to 200 hPa). In addition, the models exhibit distinct behaviors in predicting the two sub-grid scale variables: for Paraformer, the R\textsuperscript{2} values for $dT/dt$ are generally higher than for $dq/dt$ in the upper atmosphere, whereas the MLP model displays similar R\textsuperscript{2} values for both. Figures \ref{Fig:map_level1} and \ref{Fig:map_level2} further highlight the spatial variations in R\textsuperscript{2} improvements achieved by the Paraformer model. The R\textsuperscript{2} values for $dT/dt$ are significantly higher than those for $dq/dt$ at both low and high atmospheric levels, while the difference is less clear at intermediate levels.

\begin{figure}[htbp]
    \centering
    \includegraphics[scale=0.5]{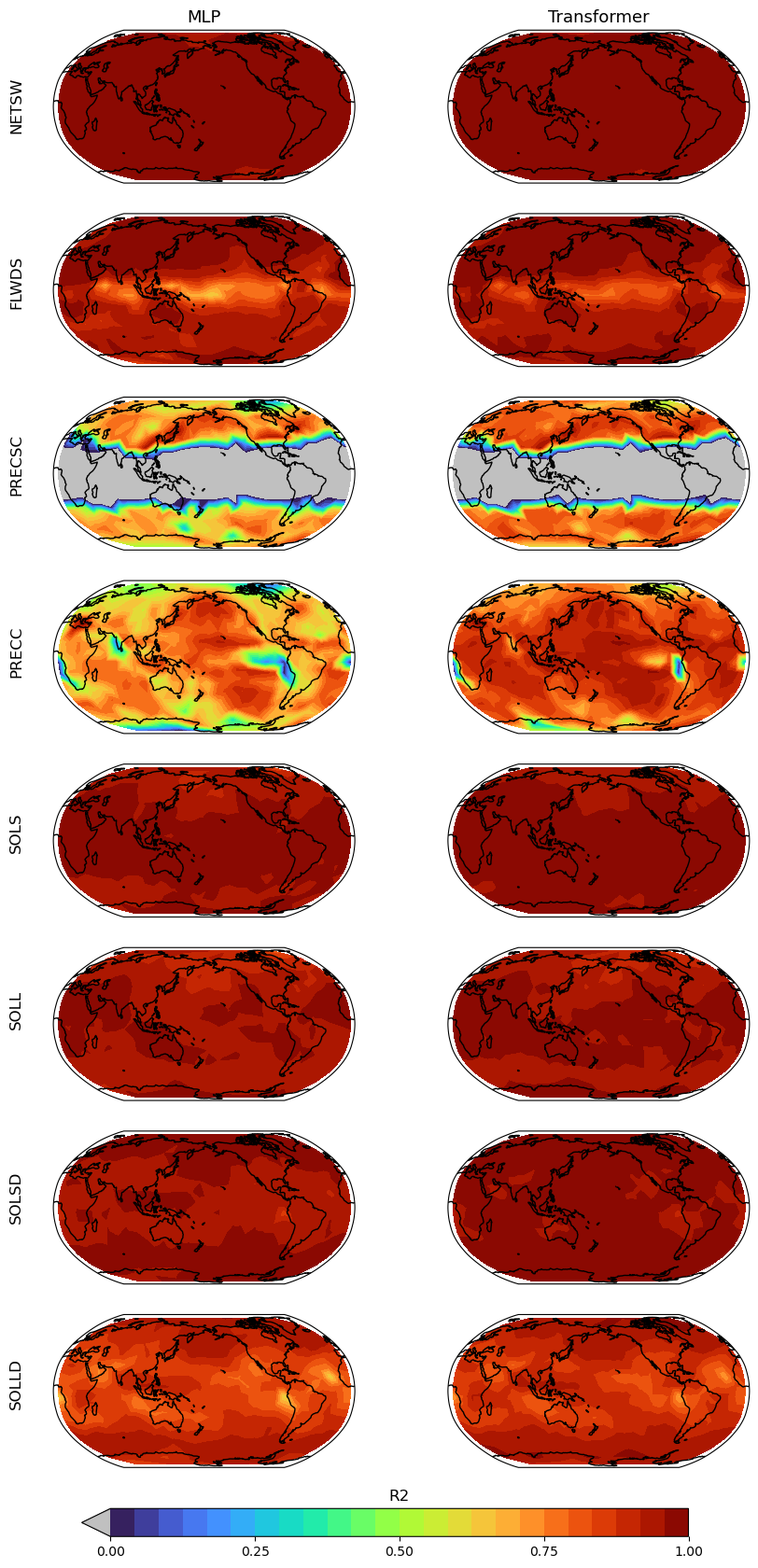}
\caption{Spatial distribution of R\textsuperscript{2} using MLP (left column) and Paraformer (right column) of 8 scalar target variables in v1. The names of the variables are labeled on the left.}
\label{Fig:3}
\end{figure}

Figure \ref{Fig:3} shows the spatial distribution of R\textsuperscript{2} values for variables without vertical structures. Among the eight scalar variables analyzed, six are related to radiative transfer processes. For NETSW, both models achieve high accuracy across all regions with R\textsuperscript{2} values approaching 1. However, for longwave fluxes such as FLWDS, R\textsuperscript{2} values are generally lower in equatorial regions, particularly over the Indian Ocean and western Pacific Ocean, where they range between 0.5 and 0.75 For other solar flux variables, the models perform better in predicting direct fluxes compared to diffused fluxes. Additionally, visible solar fluxes show higher R\textsuperscript{2} than Near-IR solar fluxes. The remaining two scalar variables relate to precipitation rates. For PRECSC, most tropical regions exhibit negative R\textsuperscript{2} values, indicating poor simulation performance. This may be attributed to an error in the configuration of the GCM in the data generation process, as snowfall is exceedingly rare in these areas. For PRECC, the Paraformer model increases the global R\textsuperscript{2} with a value of around 0.2 but produces a limited improvement in regions such as the western coast of South America. 

Figure \ref{scatter1} compares predictions to ground truth for each grid and time point for these variables. Compared to the MLP model, the Paraformer model reduces the error in outlier data points far from the reference line (representing perfect predictions), resulting in flatter scatter plots. While both models produce reasonable predictions for most variables, significant spatial-temporal variability in sub-grid scale processes, particularly those associated with solar fluxes, introduces inaccuracies. In addition, scalar variables such as radiative fluxes and precipitation have theoretical minimum values of zero, reflecting the absence of radiation or precipitation. Although these physical constraints were obeyed during data generation, both DL models occasionally violated these rules as we found small negative values in the prediction. We argue that future models should incorporate additional physical constraints to address this issue and more details will be discussed in the next chapter.

\section{Discussion}
Our analysis in this paper is based on two sets of input and target variables from the ClimSim dataset. While state-of-the-art GCMs simulate increasingly complex multi-scale physical processes, leveraging additional variables in a DL framework could further enhance simulation accuracy \citep{yu2024climsim}. A recent study utilizing a UNet architecture on ClimSim demonstrated improvements by incorporating auxiliary input variables \citep{hu2024stable}. Future research may also explore the high-resolution version of ClimSim with consideration of the associated high computational costs. 

Developing a DL framework for two-scale physical processes using datasets like ClimSim represents a critical first step toward advancing parameterization schemes. Ultimately, these architectures aim to be integrated into real-geography GCMs, replacing traditional schemes, an application called "online testing". In this context, considerations such as model size, computational cost, and training and inference time become essential. The quadratic computational complexity of the attention mechanism in the vanilla Transformer model used in this work may pose a challenge to this final step and should be examined in future research. In recent years, several strategies have emerged to reduce Transformer complexity \citep{shen2021efficient,kitaev2020reformer}, and applying these architectures to parameterization problems could significantly alleviate computational burdens in the future. 

Currently, our Transformer model focuses solely on the time dimension of the data, assuming no spatial correlation between grids in GCMs. Future work could address this limitation by testing Transformer architectures designed for spatiotemporal data, such as EarthFormer \citep{gao2022earthformer}, which might improve prediction accuracy. However, as previously noted, such architectures may not be directly applicable to ClimSim due to the uneven spatial distribution of grids.  

Additionally, incorporating physical constraints into DL models offers a promising direction. Physics-Informed Neural Networks (PINNs), which have been successfully applied to weather and climate data analysis \citep{kashinath2021physics}, could further enhance model performance by ensuring adherence to the physical principles governing sub-grid scale processes. This approach may not only enhance global predictions but also provide deeper insights into the complex dynamics of upper-level sub-grid variables.

\section{Conclusion}
In this paper, we propose a "memory-aware" Transformer-based climate parameterization scheme and evaluate its performance on two variable sets from the ClimSim dataset. Our results demonstrate that by sequencing data along the temporal dimension, the Transformer's attention mechanism effectively captures changes in sub-grid scale physical processes in the Earth system, outperforming most classical DL architectures. The prediction accuracy and the extent of improvements vary across regions and atmospheric levels. Additionally, increasing the dataset size enhances the accuracy of sub-grid process predictions. This work bridges the gap in applying advanced attention mechanisms to climate parameterization and offers valuable insights for developing future DL-based parameterization schemes.

\section*{Acknowledgment}
This work is supported by the U.S. Department of Defense (DoD) Strategic Environmental Research and Development Program (SERDP) (\#RC20-1183) and the Indian Monsoon Mission project (\#IITM/MM-III/IND-4).

\newpage
\bibliography{tackling_climate_workshop.bib}

\clearpage

\appendix\label{appendix_a}
\section{Supplementary Information}
\setcounter{table}{0}
\renewcommand{\thetable}{\Alph{section}\arabic{table}}
\renewcommand*{\theHtable}{\thetable}
\begin{table}[htbp]
\centering
\caption{Input and target variables in v1 in the low-resolution, real-geography ClimSim dataset. Some variables have a size of 60, corresponding to the number of vertical levels in the atmosphere, others are scalar variables.}
\label{tab:v1}
\resizebox{\textwidth}{!}{%
\begin{tabular}{@{}llll@{}}
\toprule
\textbf{Input (grid-scale variables)} & Size & \textbf{Output (sub-grid variables)}     & Size \\ \midrule
Temperature {[}K{]}                   & 60 & Heating tendency, $dT/dt$ {[}K/s{]}                & 60 \\
Specific humidity {[}kg/kg{]}         & 60   & Moistening tendency, $dq/dt$ {[}kg/kg/s{]} & 60   \\
Surface pressure {[}Pa{]}             & 1  & Net surface shortwave flux, NETSW {[}W/m\textsuperscript{2}{]}     & 1  \\
Insolation {[}W/m\textsuperscript{2}{]}                 & 1  & Downward surface longwave flux, FLWDS {[}W/m\textsuperscript{2}{]} & 1  \\
Surface latent heat flux {[}W/m\textsuperscript{2}{]}   & 1  & Snow rate, PRECSC {[}m/s{]}                      & 1  \\
Surface sensible heat flux {[}W/m\textsuperscript{2}{]} & 1  & Rain rate, PRECC {[}m/s{]}                       & 1  \\
                                      &    & Visible direct solar flux, SOLS {[}W/m\textsuperscript{2}{]}       & 1  \\
                                      &    & Near-IR direct solar flux, SOLL {[}W/m\textsuperscript{2}{]}       & 1  \\
                                      &    & Visible diffused solar flux, SOLSD {[}W/m\textsuperscript{2}{]}    & 1  \\
                                      &    & Near-IR diffused solar flux, SOLLD {[}W/m\textsuperscript{2}{]}    & 1  \\
\midrule                                      
\textbf{Total}                                  & 124  & \textbf{Total}                                              & 128  \\
\bottomrule
\end{tabular}%
}
\end{table}

% \appendix\label{appendix_b}
% \section{Summary of Variables}
% \setcounter{table}{0}
% \renewcommand{\thetable}{\Alph{section}\arabic{table}}
% \renewcommand*{\theHtable}{\thetable}

% \usepackage{graphicx}

% \usepackage{booktabs}
% \usepackage{graphicx}
\begin{table}[htbp]
\caption{Same as Table \ref{tab:v1}, but for v2.}
\label{tab:v2}
\resizebox{\textwidth}{!}{%
\begin{tabular}{@{}llll@{}}
\toprule
\textbf{Input (grid-scale variables)}           & Size & \textbf{Output (sub-grid scale variables)}                  & Size \\ \midrule
Temperature {[}K{]}                             & 60   & Heating tendency, $dT/dt$ {[}K/s{]}                         & 60   \\
Specific humidity {[}kg/kg{]}                   & 60   & Moistening tendency, $dq/dt$ {[}kg/kg/s{]}                  & 60   \\
Cloud liquid mixing ratio {[}kg/kg{]}           & 60   & Liquid cloud tendency, $dq_l/dt$ {[}kg/kg/s{]}           & 60   \\
Cloud ice mixing ratio {[}kg/kg{]}              & 60   & Ice cloud tendency, $dq_i/dt$ {[}kg/kg/s{]}              & 60   \\
Zonal wind speed {[}m/s{]}                      & 60   & Zonal wind tendency, $du/dt$ {[}m/s\textsuperscript{2}{]}                             & 60   \\
Meridional wind speed {[}m/s{]}                 & 60   & Meridional wind tendency, $dv/dt$ {[}m/s\textsuperscript{2}{]}                        & 60   \\
Ozone volume mixing ratio {[}mol/mol{]}         & 60   & Net surface shortwave flux, NETSW [W/m\textsuperscript{2}] & 1    \\
Methane volume mixing ratio {[}mol/mol{]}       & 60   & Downward surface longwave flux, FLWDS {[}W/m\textsuperscript{2}{]}            & 1    \\
Nitrous oxide volume mixing ratio {[}mol/mol{]} & 60   & Snow rate, PRECSC {[}m/s{]}                                 & 1    \\
Surface pressure {[}Pa{]}                       & 1    & Rain rate, PRECC {[}m/s{]}                                  & 1    \\
Insolation {[}W/m\textsuperscript{2}{]}                           & 1    & Visible direct solar flux, SOLS {[}W/m\textsuperscript{2}{]}                  & 1    \\
Surface latent heat flux {[}W/m\textsuperscript{2}{]}             & 1    & Near-IR direct solar flux, SOLL {[}W/m\textsuperscript{2}{]}                  & 1    \\
Surface sensible heat flux {[}W/m\textsuperscript{2}{]}           & 1    & Visible diffused solar flux, SOLSD {[}W/m\textsuperscript{2}{]}               & 1    \\
Zonal surface stress {[}W/m\textsuperscript{2}{]}                 & 1    & Near-IR diffused solar flux, SOLLD {[}W/m\textsuperscript{2}{]}               & 1    \\
Meridional surface stress {[}W/m\textsuperscript{2}{]}            & 1    &                                                             &      \\
Cosine of solar zenith angle                    & 1    &                                                             &      \\
Albedo for diffuse longwave radiation           & 1    &                                                             &      \\
Albedo for direct longwave radiation            & 1    &                                                             &      \\
Albedo for diffuse shortwave radiation          & 1    &                                                             &      \\
Albedo for direct shortwave radiation           & 1    &                                                             &      \\
Upward longwave flux {[}W/m\textsuperscript{2}{]}                 & 1    &                                                             &      \\
Sea-ice area fraction                           & 1    &                                                             &      \\
Land area fraction                              & 1    &                                                             &      \\
Ocean area fraction                             & 1    &                                                             &      \\
Snow depth over ice                             & 1    &                                                             &      \\
Snow depth over land (liquid water equivalent)  & 1    &                                                             &      \\
\midrule
\textbf{Total}                                  & 557  & \textbf{Total}                                              & 368  \\ \bottomrule
\end{tabular}%
}
\end{table}

% \section{Supplementary Figures}
\renewcommand{\thefigure}{A\arabic{figure}}
\setcounter{figure}{0}
\begin{figure}[htbp]
    \centering
    \includegraphics[width=1\linewidth]{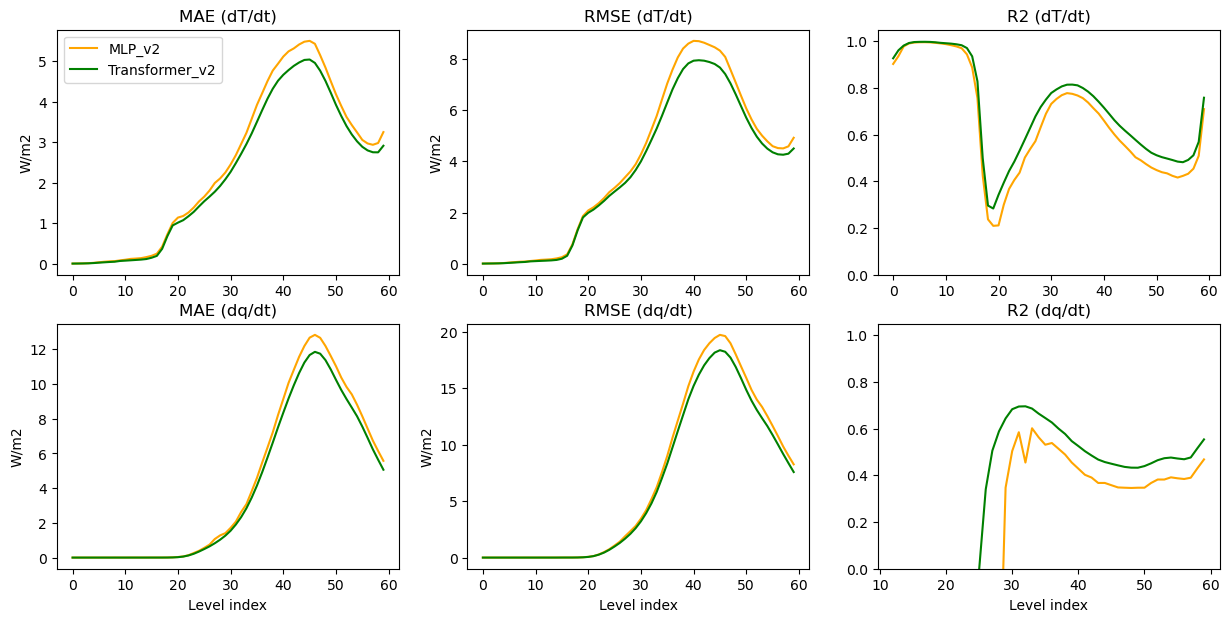}
    \caption{Same as Figure \ref{fig:1}, but for v2.}
    \label{fig1 v2}
\end{figure}

% \section{Supplementary Figures}
\renewcommand{\thefigure}{A\arabic{figure}}
\begin{figure}[htbp]
    \centering
    \includegraphics[width=1\linewidth]{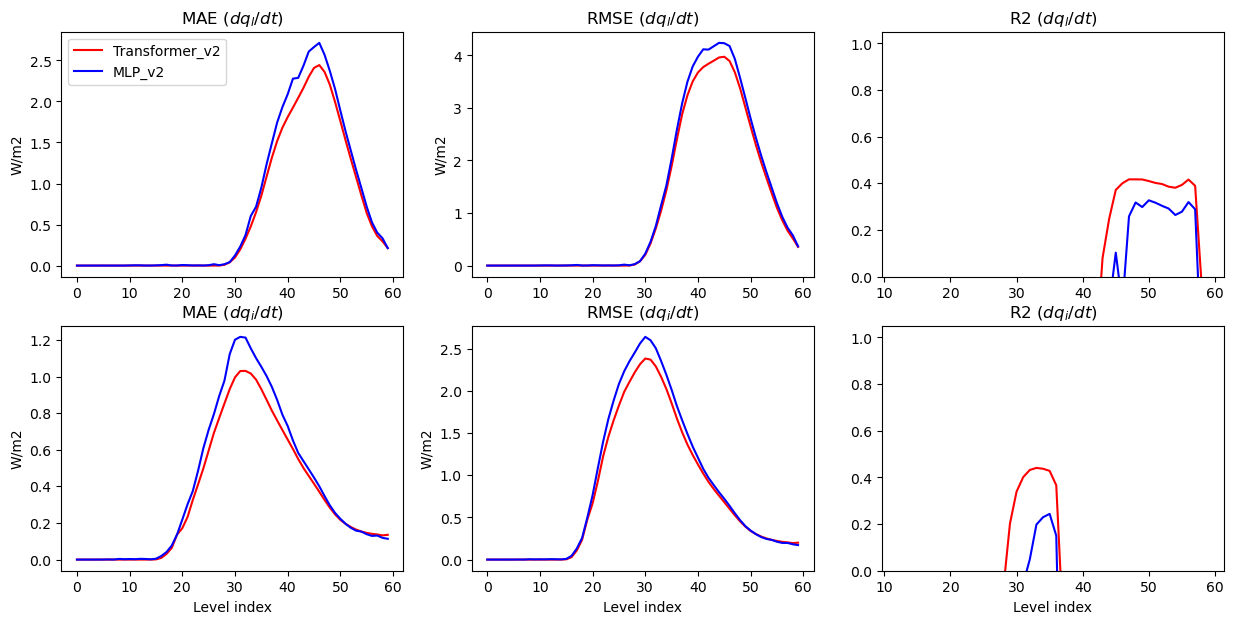}
    \caption{Same as Figure \ref{fig:1}, but for two other variables in v2: liquid cloud tendency $dq_l/dt$ and ice cloud tendency $dq_i/dt$.}
    \label{fig1 v2 cloud}
\end{figure}

% \usepackage{booktabs}
% \usepackage{multirow}
% \usepackage{graphicx}
\begin{table}[htbp]
\caption{MAE, RMSE and R\textsuperscript{2} of MLP and Paraformer on the variable set v2. Negative R\textsuperscript{2} values are not shown.}
\label{tab:metric v2}
\resizebox{\textwidth}{!}{%
\begin{tabular}{@{}l|lll|lll@{}}
\toprule
\multicolumn{1}{c|}{\multirow{2}{*}{\textbf{Variables}}} & \multicolumn{3}{c|}{\textbf{MLP}}          & \multicolumn{3}{c}{\textbf{Paraformer}}   \\ \cmidrule(l){2-7} 
\multicolumn{1}{c|}{}                                    & \textbf{MAE} & \textbf{RMSE} & \textbf{R\textsuperscript{2}} & \textbf{MAE} & \textbf{RMSE} & \textbf{R\textsuperscript{2}} \\ \midrule
$dT/dt$                                                 & 2.375        & 3.860         & 0.658       & 2.180        & 3.570         & 0.708       \\
$dq/dt$                                              & 4.174        & 6.712         & -           & 3.836        & 6.236         & -           \\
$dq_l/dt$                                             & 0.715        & 1.223         & -           & 0.640        & 1.139         & -           \\
$dq_i/dt$                                             & 0.394        & 0.837         & -           & 0.348        & 0.766         & -           \\
$du/dt$                                                & 1.85E-05     & 9.83E-05      & -           & 1.80E-05     & 9.84E-05      & -           \\
$dv/dt$                                                  & 1.80E-05     & 9.42E-05      & -           & 1.74E-05     & 9.47E-05      & -           \\
NETSW                                          & 9.373        & 18.037        & 0.991       & 7.261        & 15.186        & 0.995       \\
FLWDS                                          & 4.668        & 6.045         & 0.933       & 3.806        & 5.188         & 0.953       \\
PRECSC                                         & 2.015        & 3.514         & -           & 1.539        & 2.857         & -           \\
PRECC                                          & 28.906       & 56.491        & -           & 19.062       & 41.910        & -           \\
SOLS                                           & 6.393        & 13.244        & 0.975       & 5.387        & 11.927        & 0.980       \\
SOLL                                           & 8.894        & 18.290        & 0.961       & 7.861        & 17.136        & 0.967       \\
SOLSD                                          & 3.619        & 7.155         & 0.975       & 2.970        & 6.279         & 0.980       \\
SOLLD                                          & 4.232        & 8.693         & 0.902       & 3.806        & 8.177         & 0.913       \\ \bottomrule
\end{tabular}%
}
\end{table}

\renewcommand{\thefigure}{A\arabic{figure}}
\begin{figure}[htbp]
   \begin{minipage}{0.48\textwidth}
     \centering
     \includegraphics[width=1\linewidth]{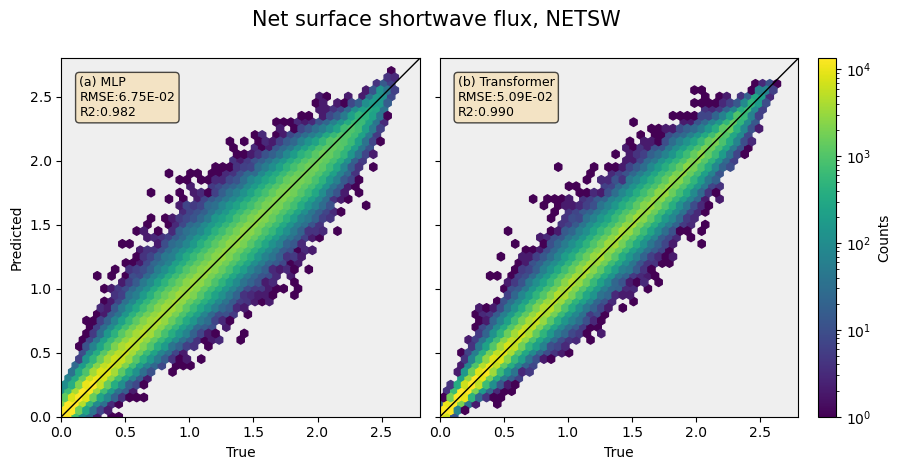}
   \end{minipage}\hfill
   \begin{minipage}{0.48\textwidth}
     \centering
     \includegraphics[width=1\linewidth]{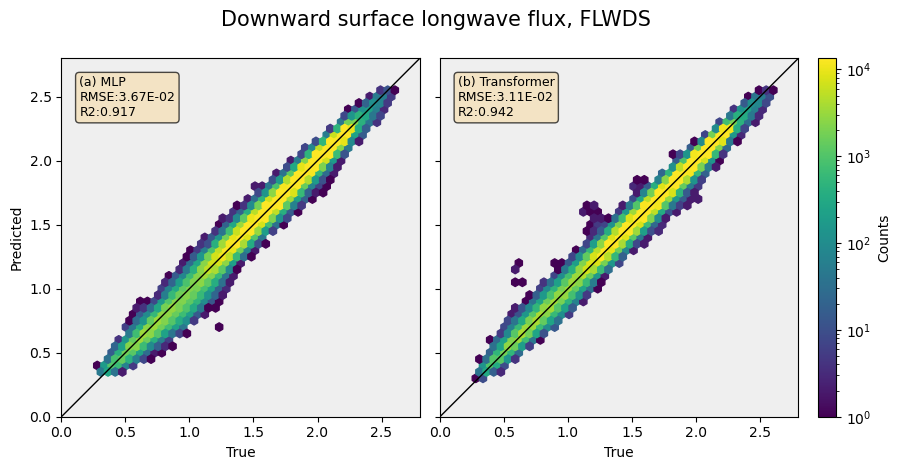}
   \end{minipage}
    \begin{minipage}{0.48\textwidth}
     \centering
     \includegraphics[width=1\linewidth]{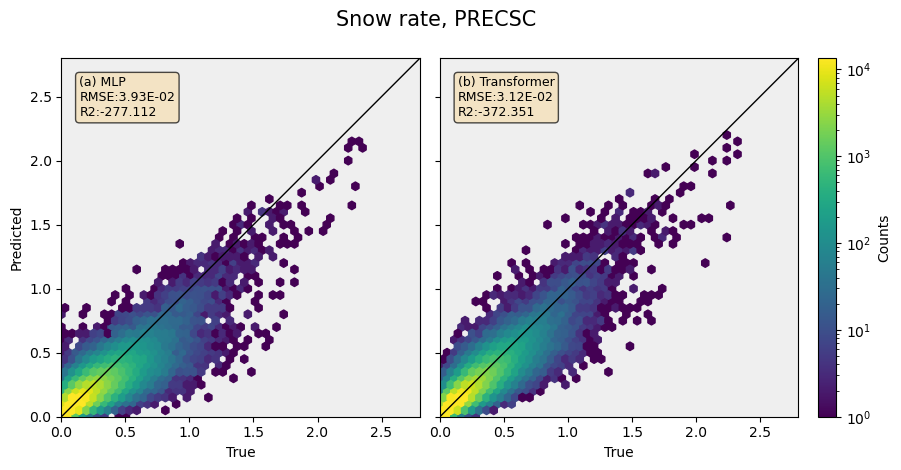}
   \end{minipage}\hfill
   \begin{minipage}{0.48\textwidth}
     \centering
     \includegraphics[width=1\linewidth]{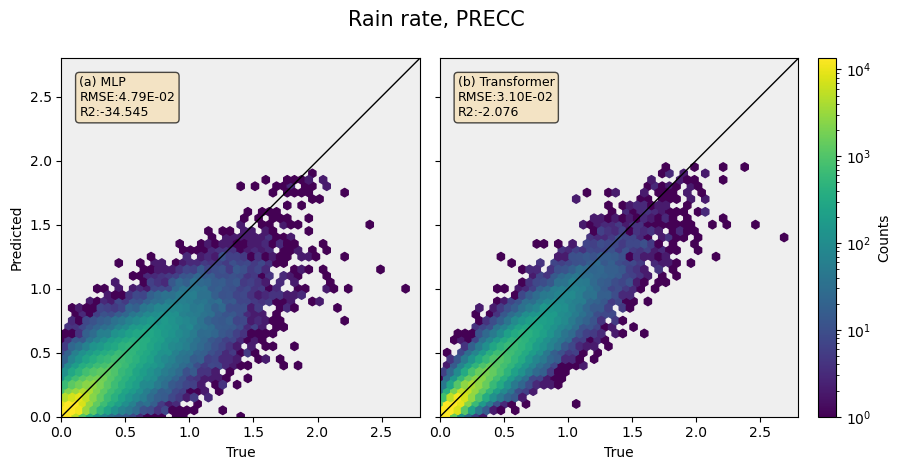}
   \end{minipage}\hfill
   \begin{minipage}{0.48\textwidth}
     \centering
     \includegraphics[width=1\linewidth]{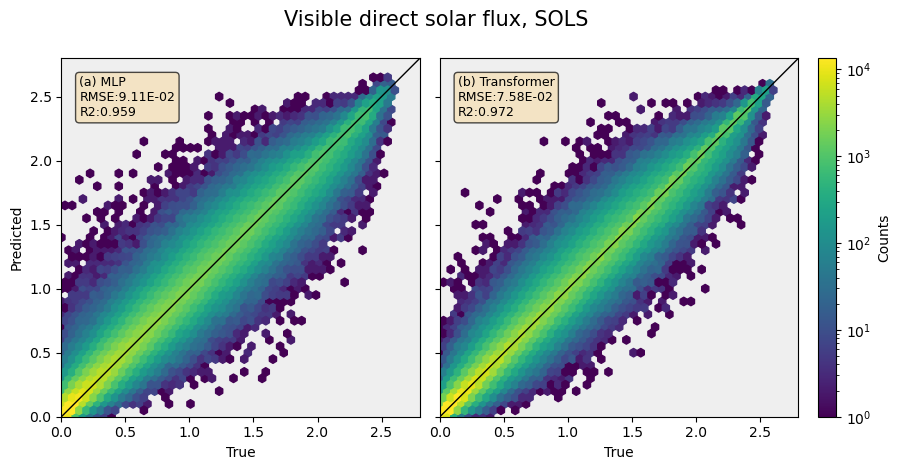}
   \end{minipage}\hfill
    \begin{minipage}{0.48\textwidth}
     \centering
     \includegraphics[width=1\linewidth]{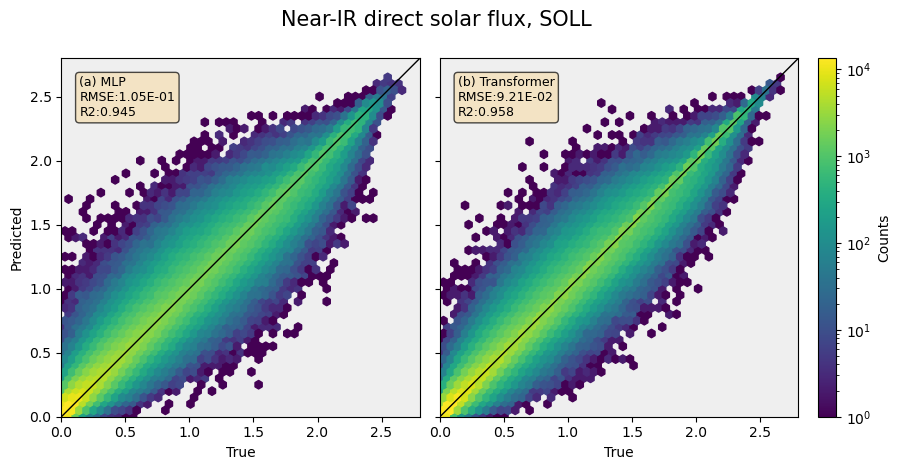}
   \end{minipage}\hfill
   \begin{minipage}{0.48\textwidth}
     \centering
     \includegraphics[width=1\linewidth]{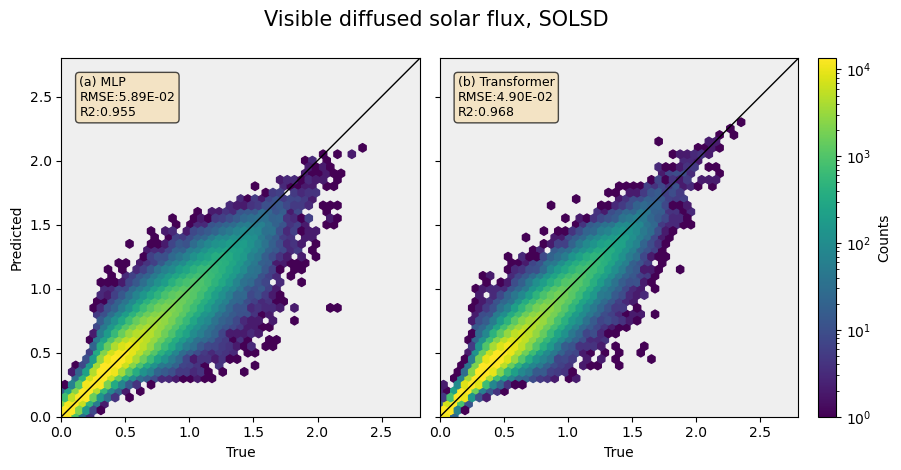}
   \end{minipage}\hfill
   \begin{minipage}{0.48\textwidth}
     \centering
     \includegraphics[width=1\linewidth]{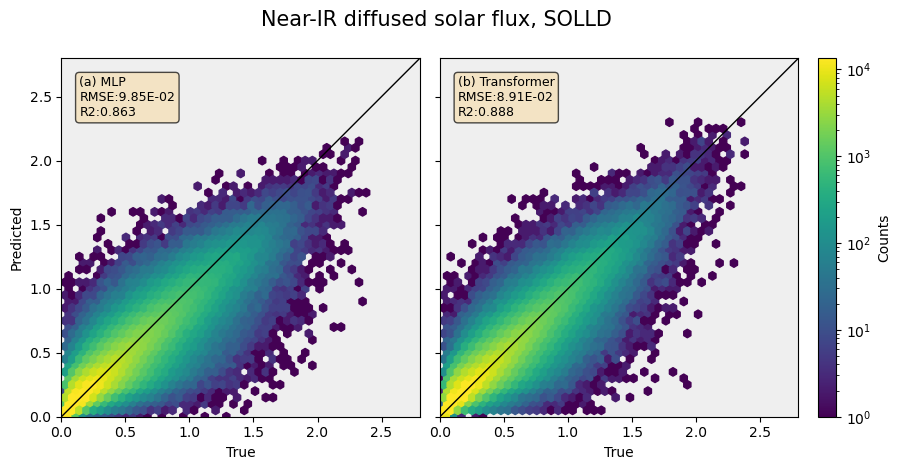}
   \end{minipage}\hfill

\caption{Ground truth versus predictions from MLP and Paraformer of 8 scalar variables in v1. A 45° reference line indicating no-error prediction is shown in each figure. The color of each hexagon indicates the number of data points it encloses.}
\label{scatter1}
\end{figure}

\renewcommand{\thefigure}{A\arabic{figure}}
\begin{figure}[htbp]
   \begin{minipage}{0.48\textwidth}
     \centering
     \includegraphics[width=1\linewidth]{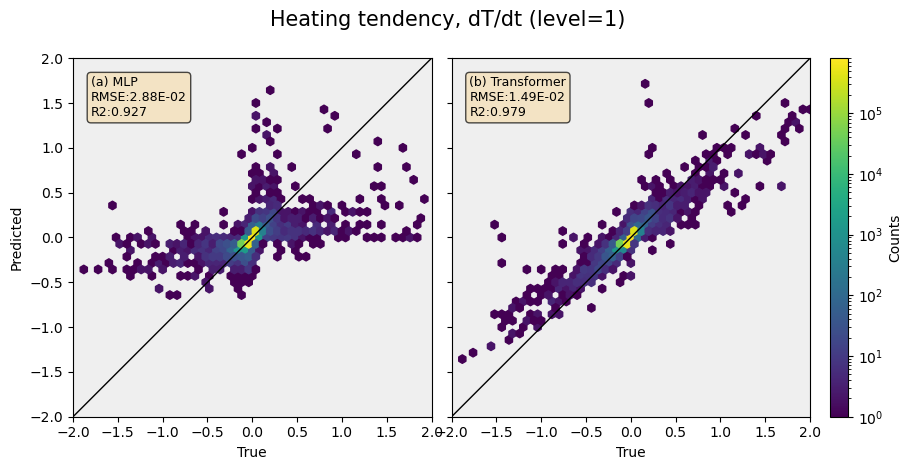}
   \end{minipage}\hfill
   \begin{minipage}{0.48\textwidth}
     \centering
     \includegraphics[width=1\linewidth]{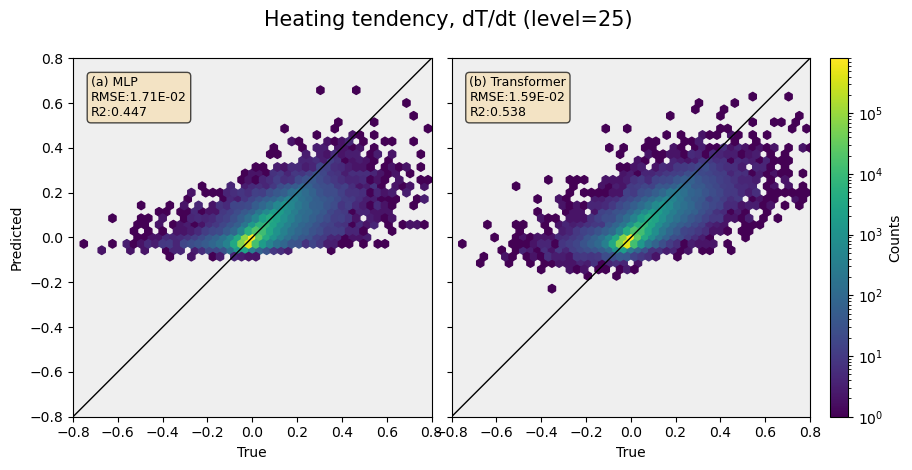}
   \end{minipage}
    \begin{minipage}{0.48\textwidth}
     \centering
     \includegraphics[width=1\linewidth]{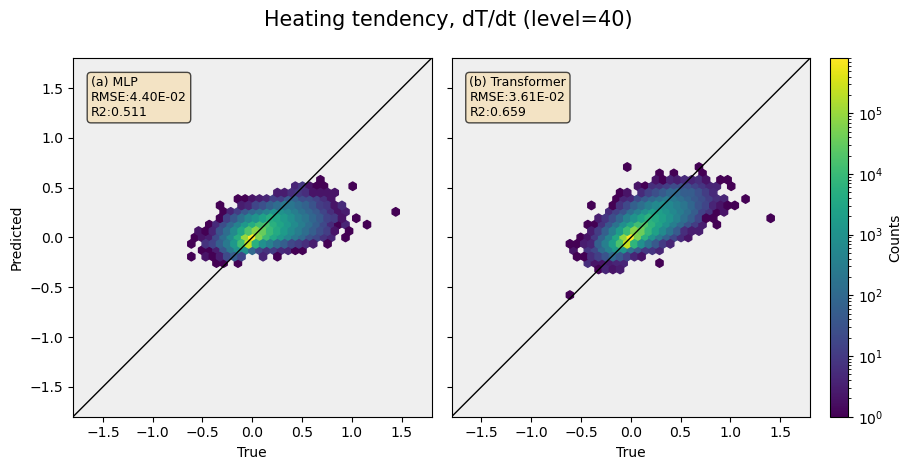}
   \end{minipage}\hfill
   \begin{minipage}{0.48\textwidth}
     \centering
     \includegraphics[width=1\linewidth]{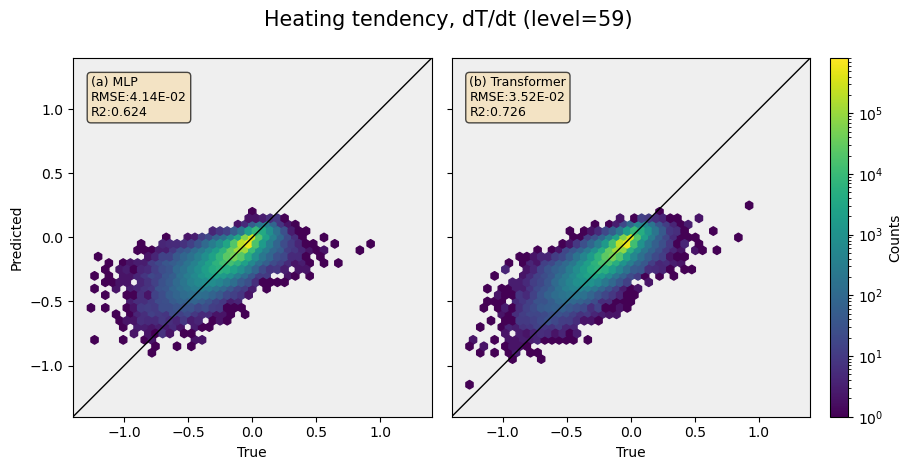}
   \end{minipage}\hfill
\caption{Same as Figure \ref{scatter1}, but for $dT/dt$ at 4 atmospheric levels (1, 25, 40, 59) in v1.}
\label{scatter2}
\end{figure}

\renewcommand{\thefigure}{A\arabic{figure}}
\begin{figure}[htbp]
   \begin{minipage}{0.48\textwidth}
     \centering
     \includegraphics[width=1\linewidth]{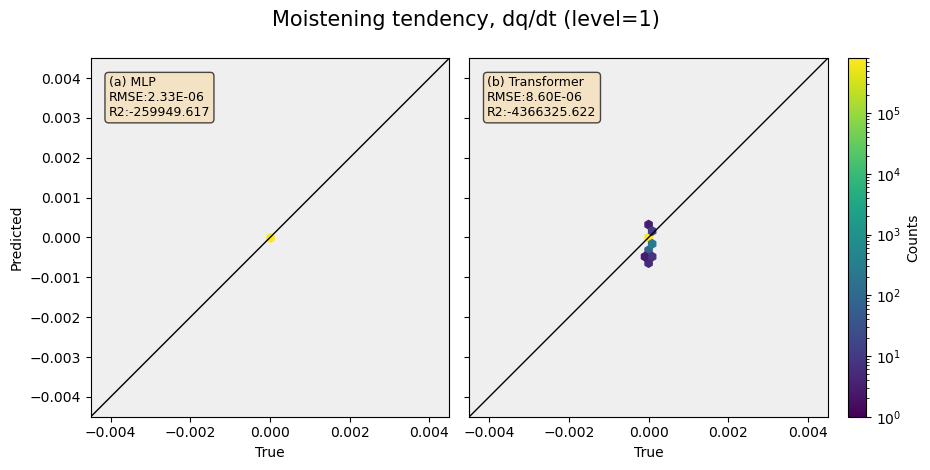}
   \end{minipage}\hfill
   \begin{minipage}{0.48\textwidth}
     \centering
     \includegraphics[width=1\linewidth]{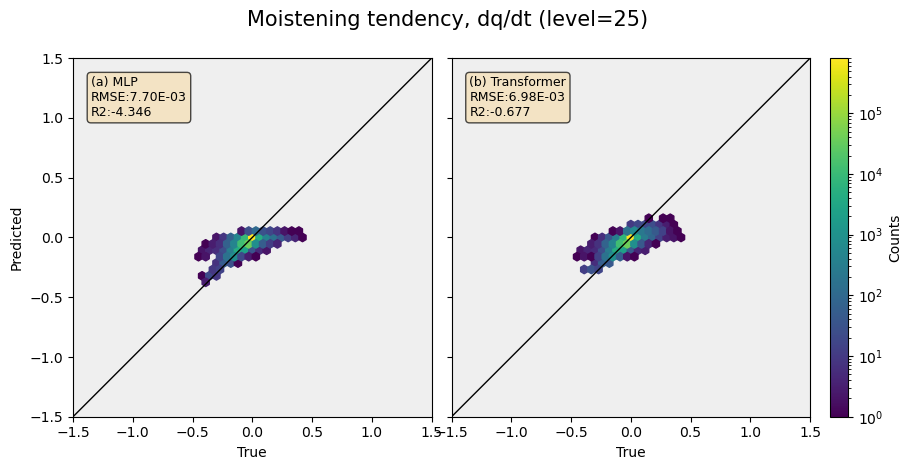}
   \end{minipage}\hfill
   \begin{minipage}{0.48\textwidth}
     \centering
     \includegraphics[width=1\linewidth]{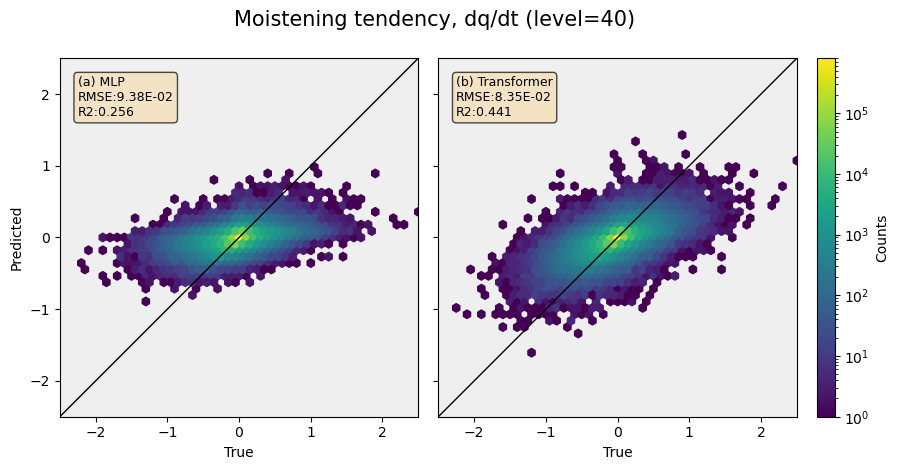}
   \end{minipage}\hfill
   \begin{minipage}{0.48\textwidth}
     \centering
     \includegraphics[width=1\linewidth]{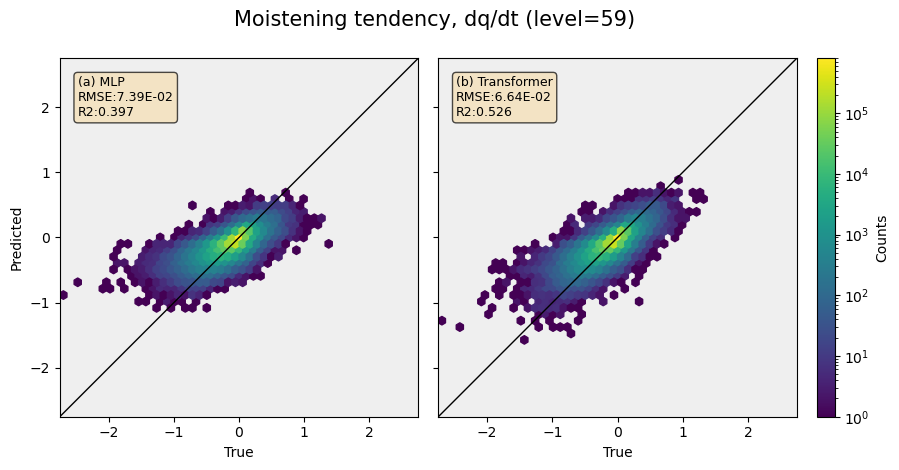}
   \end{minipage}
\caption{Same as Figure \ref{scatter1}, but for $dq/dt$ at 4 atmospheric levels (1, 25, 40, 59) in v1.}
\label{scatter3}
\end{figure}

\renewcommand{\thefigure}{A\arabic{figure}}
\begin{figure}[htbp]
\centering
\includegraphics[scale=0.6]{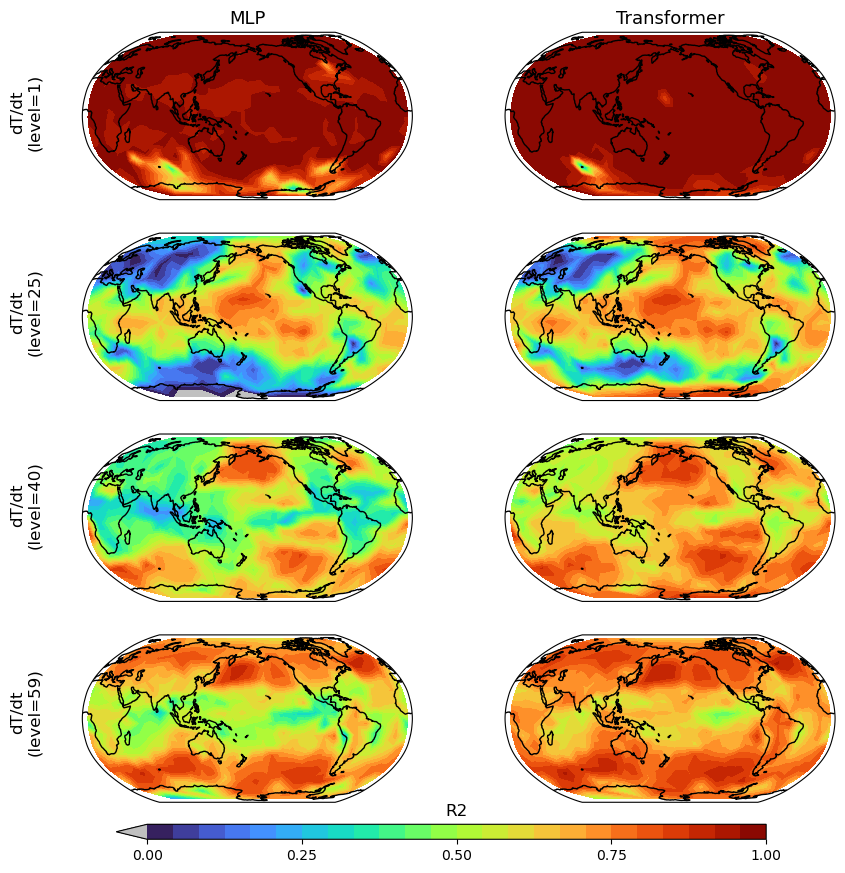}
   % \begin{minipage}{0.48\textwidth}
   %   \centering
   %   \includegraphics[width=1\linewidth]{map_level1.png}
   % \end{minipage}\hfill
   % \begin{minipage}{0.48\textwidth}
   %   \centering
   %   \includegraphics[width=1\linewidth]{map_level5.png}
   % \end{minipage}
   %  \begin{minipage}{0.48\textwidth}
   %   \centering
   %   \includegraphics[width=1\linewidth]{map_level2.png}
   % \end{minipage}\hfill
   % \begin{minipage}{0.48\textwidth}
   %   \centering
   %   \includegraphics[width=1\linewidth]{map_level6.png}
   % \end{minipage}\hfill
   % \begin{minipage}{0.48\textwidth}
   %   \centering
   %   \includegraphics[width=1\linewidth]{map_level3.png}
   % \end{minipage}\hfill
   %  \begin{minipage}{0.48\textwidth}
   %   \centering
   %   \includegraphics[width=1\linewidth]{map_level7.png}
   % \end{minipage}\hfill
   % \begin{minipage}{0.48\textwidth}
   %   \centering
   %   \includegraphics[width=1\linewidth]{map_level4.png}
   % \end{minipage}\hfill
   % \begin{minipage}{0.48\textwidth}
   %   \centering
   %   \includegraphics[width=1\linewidth]{map_level8.png}
   % \end{minipage}\hfill

\caption{Same as Figure \ref{Fig:3}, but for $dT/dt$ at 4 atmospheric levels (1, 25, 40, 59) in v1.}
\label{Fig:map_level1}
\end{figure}

\renewcommand{\thefigure}{A\arabic{figure}}
\begin{figure}[htbp]
\centering
\includegraphics[scale=0.6]{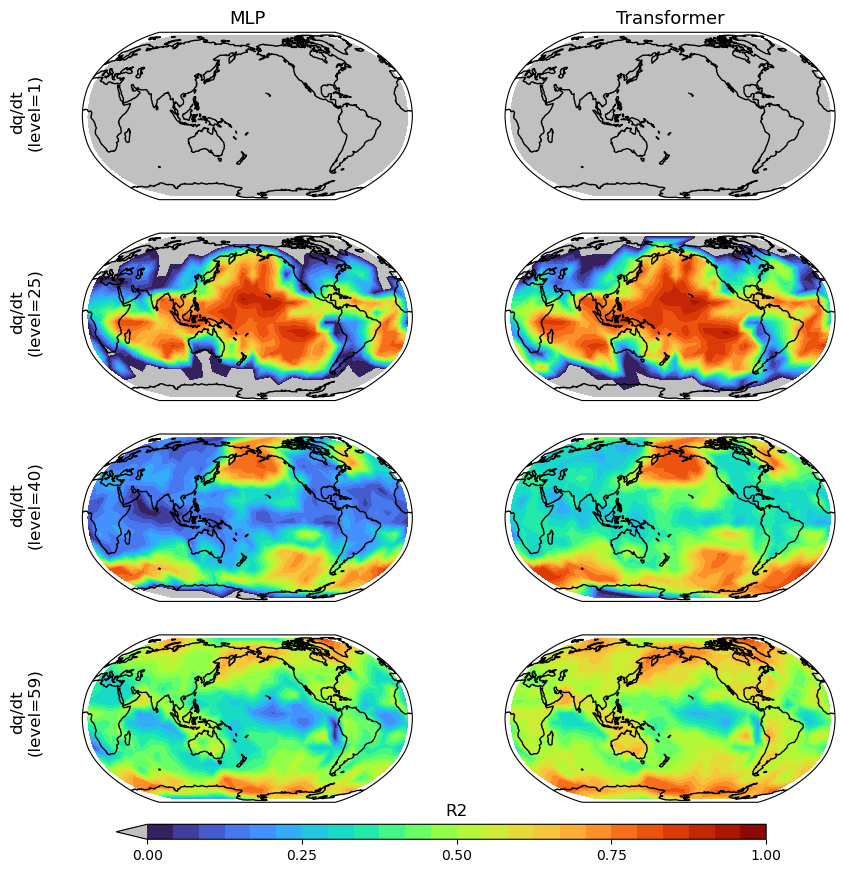}
\caption{Same as Figure \ref{Fig:3}, but for $dq/dt$ at 4 atmospheric levels (1, 25, 40, 59) in v1.}
\label{Fig:map_level2}
\end{figure}

\end{document}